\def\year{2019}\relax
\documentclass[letterpaper]{article} 
\usepackage{aaai19}  
\usepackage{times}  
\usepackage{helvet}  
\usepackage{courier}  
\usepackage{url}  
\usepackage{graphicx}  
\usepackage{amsfonts,amssymb,amsmath} 
\usepackage{algorithm}
\usepackage{algorithmic}
\usepackage{setspace}
\usepackage{graphicx}
\usepackage{booktabs}
\usepackage{multirow}
\usepackage{multicol}
\usepackage{comment}
\usepackage{subfigure}
\usepackage{balance}
\usepackage{array}
\usepackage[marginal]{footmisc}
\begin{document}
%
\title{Improve Diverse Text Generation by Self Labeling \\
Conditional Variational Auto Encoder}
 \author{Yuchi Zhang \thanks{Correspondence to Yuchi Zhang(yuchi.zyc@alibaba-inc.com)}, Yongliang Wang, Liping Zhang, Zhiqiang Zhang, Kun Gai\\
 Alibaba Group\\
 Beijing, China, 100102\\
 }
\maketitle
\begin{abstract}
Diversity plays a vital role in many text generating applications.
In recent years, Conditional Variational Auto Encoders (CVAE) have shown promising performances for this task.
However, they often encounter the so called KL-Vanishing problem.
Previous works mitigated such problem by heuristic methods such as strengthening the encoder or weakening the decoder
while optimizing the CVAE objective function.
Nevertheless, the optimizing direction of these methods are implicit and it is hard to find an appropriate degree to which these methods should be applied.
In this paper, we propose an explicit optimizing objective to complement the CVAE to directly pull away from KL-vanishing.
In fact, this objective term guides the encoder towards the ``best encoder" of the decoder to enhance the expressiveness.
A labeling network is introduced to estimate the ``best encoder".
It provides a continuous label in the latent space of CVAE to help build a close connection between latent variables and targets.
The whole proposed method is named Self Labeling CVAE~(SLCVAE).
To accelerate the research of diverse text generation, we also propose a large native \emph{one-to-many} dataset.
Extensive experiments are conducted on two tasks, which show that our method largely improves the generating diversity while achieving comparable accuracy compared with state-of-art algorithms.
\end{abstract}

\section{Introduction}
Text generating techniques are widely used in various tasks, such as dialogue generation~\cite{serban2016building,li2016deep,zhao2017learning}, image caption~\cite{chen2015mind,vinyals2015show,xu2015show} and question-answer systems~\cite{beamon2017justifying,oh2017non}, etc.
Recently, encoder-decoder models such as SEQ2SEQ\cite{sutskever2014sequence} have been increasingly adopted in text generating tasks.
Encoder-decoder models extract a semantic representation from the input and generate sentences coherent to the input according to this representation.
They perform well in tasks which require accuracy and relativeness.
However, applications such as dialogue systems further require results with diversity besides accuracy.
Conventional encoder-decoder models are not good at handling such situations due to its deterministic nature.

\begin{figure}[t]
	\centering  
	\includegraphics[width=0.95\linewidth]{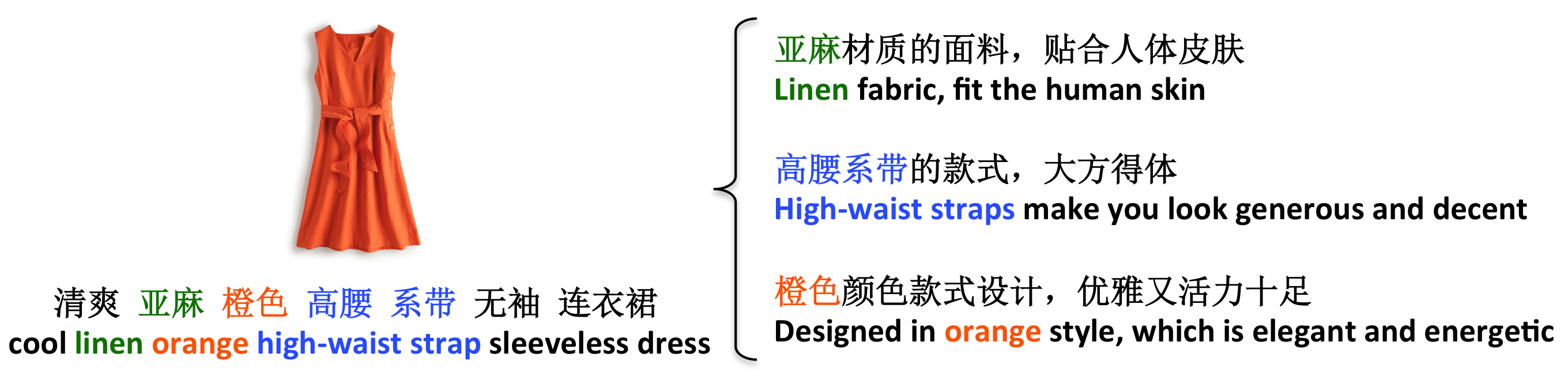}  
	\caption{Example of recommendation texts of a dress. Some of them emphasize on the linen material  while some others emphasize on the color etc.}  
	\label{fig:one-to-many-example}   
\end{figure}

In open domain conversation systems, given the dialogue history, there may exist various kinds of responses which are grammatically correct and semantically meaningful.
The dialogue bots should be able to model these multiple responses for the same input in training and give diverse answers like humans in predicting.
Another application takes place in e-commerce recommendation systems.
For a given item, multiple selling points and descriptions are needed for personalized recommendations. 
Fig.~\ref{fig:one-to-many-example} shows an example.
With plenty of recommendation texts, different sentences can be selected to display to different users to meet their preferences or to the same user at various situations so that he/she does not feel monotonous.

We summarize the above applications as the ``one source, multiple targets'' problems.
As conventional encoder-decoder models encode same input patterns to same unique representative vectors without any variation, their ability of generating different sentences from one input are limited. 

Researchers have made efforts to improve the encoder-decoder models for more diverse generations.
In the early periods, methods are proposed to interfere the inference stage of a well-trained encoder-decoder model to encourage abundant outputs~\cite{li2016diversity,vijayakumar2016diverse}.
The drawback of such methods is that they do not optimize the encoder-decoder models to fit multi-target data and the quality of their generating results is limited by the trade-off between accuracy and diversity.
Recently, variational encoder-decoders have shown great potentials in solving the ``one source, multiple targets" problems~\cite{bowman2016generating,zhao2017learning,shen2018improving}.
These methods introduced an intermediate latent variable and assume that each configuration of the latent variable corresponds to a feasible response.
Thus diverse responses can be generated by sampling the variable.
However, both VAE and CVAE have encountered the KL-vanishing problem that the decoder tends to model the targets without making use of the latent variables.

In this paper, we point out that during optimizing the objective of CVAE, the encoder is gradually pulled to a prior distribution and losing discriminative ability of the targets, while the decoder tends to fit the data even without the help of encoders.
Thus KL-vanishing is rooted in the objective of CVAE.
Current approaches, either weakening the decoder or strengthening the encoder to make compensation to the objective implicitly in advance, only mitigate this problem and are hard to determine how weak/strong should decoder/encoder be. 
Orthogonal to these efforts, we propose an explicit optimization objective for the encoder to move towards better expressiveness to fit current decoder.
With this novel objective, the latent variable distribution from the encoder has the potential to be appropriately flexible in correspondence with decoder, which naturally coordinates the representative abilities of the encoder and decoder and enhances the utilization of latent variable by decoders.
Specifically, an additional module called ``labeling network'' is used to estimate the ``best encoder'' for the current decoder.
Then a loss which measures the difference between the latent variable of CVAE and predicted variable from labeling network is added to the original objective function of the CVAE.
Since this loss pulls the encoder towards the ``best encoder'' approximated by the labeling network and in the meanwhile original CVAE pulls encoder to the prior, an equilibrium will be reached where KL-vanishing can be avoided.
Additionally, the labeling network introduces a continuous label for each target which essentially reflects the structural constraints of the latent space.
Therefore, it guarantees each $z$ in the latent space corresponds to a unique target, thus improves the coverage of the generations in target space.
We alternatively train the ``labeling network" and the CVAE structure, and call this model Self Labeling Conditional Variational Auto Encoder (SLCVAE).

In summary our main contributions are:
Firstly, we point out that the current CVAE objective function tends to encounter the \emph{KL-Vanishing problem} due to the lack of explicit constraints on the connection between the latent variable and targets.
Secondly, we propose the self labeling mechanism which connects the decoder with latent variable by a novel explicit optimization objective.
With this objective, the encoder is pulled towards the ``best encoder" defined by current decoder and the prior distribution simultaneously, which leads to equilibrium at which the encoder distribution is close to the prior and also remains the expressiveness.
Thus the KL-vanishing problem can be significantly relieved.
Further, extensive experiments demonstrates that our method called SLCVAE owns better ability to model multiple targets and improves the diversity of text generation without losing accuracy. 
Thirdly, a large scale dataset called EGOODS which contains native one-to-many text data of high quality is constructed to accelerate the research of diverse text generation.

\begin{figure*}[h]
	\centering  
	\includegraphics[width=0.95\linewidth]{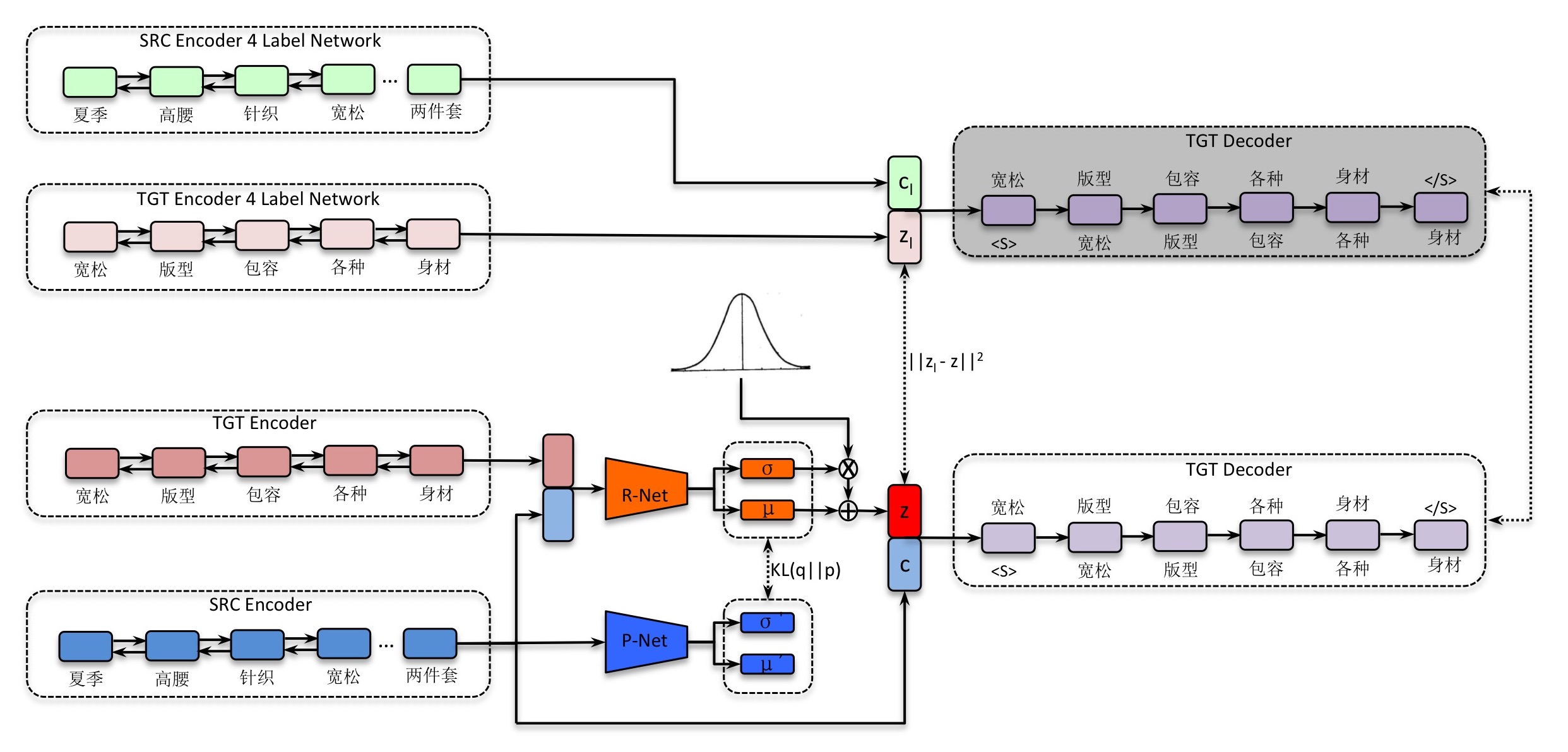}  
	\vspace{-0.2cm}
	\caption{Overview of the proposed method. The top part is the Labeling Phase and the bottom part is the CVAE Phase. The model optimized alternatively trained between the two phases. SRC and TGT are abbreviations of source and target. R-Net and P-Net are Recognition Network and Prior Network for the reparameterization trick~\cite{zhao2017learning}. $\mathcal{L}_\mathsf{re}$ denotes a reconstruction loss in ELBO and $\mathcal{L}_\mathsf{KL}$ denotes the KL divergence term.}  
	\vspace{-0.4cm}
	\label{fig:slcvae}   
\end{figure*}

\section{Related Work}
In this section, we review the development of both the encoder-decoder models and VAE/CVAE based models for text generation.

\subsection{Encoder-decoder model}
Encoder-decoder models are commonly adopted in NLP as they are able to fit complex data by end-to-end training. The presentation of SEQ2SEQ\cite{sutskever2014sequence} structure revolutionarily augmented the quality of Machine Translation (MT). And researchers soon introduce such structure into text generating systems\cite{serban2016building,vinyals2015neural}. However, the purpose of the SEQ2SEQ models is to best fit the target sequence given the source sequence. Therefore two problems might probably happen when a SEQ2SEQ model is used for generating texts. One is that SEQ2SEQ often ends up with dull and generic responses. Such situation often takes place at conversation systems because safe and meaningless responses such as ``I don't know" or ``I'm okay" have frequently appearance and then captured by the decoder.
The other is the lack of ability of fitting multiple probable outputs,for the representative vector the decoder used to generate output is fixed and only depends on the inputs. These problems not only reduce the precision, but also limit the diversity of text generation.

To tackle the above problems. Different ways of solutions have been proposed. \cite{li2016diversity} pointed out that the mutual information of sources and targets should be augmented during the decoding procedure. They proposed the MMI-antiLM algorithm which adds a language model penalty to unconditional high frequent responses. Their algorithm successfully solved the generic and dull response problem, is not applicable to handle the multi-target, since it just consider one target at a time. Beam Search (BS) methods for n-best outputs during the decoding procedure are commonly used in MT and could be introduced to text generation. However, as the greedy strategy in BS makes it tend to generate outputs with same prefixes, sentences generated respective to one source still look similar. \cite{vijayakumar2016diverse} modifies the strategy used in BS to be subject to a diverse behavior goal by reinforcement learning. However such method might reduce the coherence of the outputs. The above methods are based on improved strategies during the inference stage of a encoder-decoder model. But the encoder-decoder model itself is not substantially made better. Their strategies are actually a trade-off between diversity and coherence and thus are restricted.

Additional information could still be used such as topics, speakers' characteristics in a dialogue session or linguistic prior knowledges. These methods, however, are not applicable for common as extra inputs are required for their unique applications.

\subsection{VAE and CVAE}
Variational Auto Encoder~\cite{rezende2014stochastic,kingma2013auto} is a popular generative model. It makes use of a latent variable \begin{math}z\end{math} sampled from a prior distribution to generate data \begin{math}x\end{math}. The logarithm likelihood of the data \begin{math}x\end{math} is optimized by maximizing the evidence lower bound (ELBO):
\begin{equation}
{\log{p(x)}}\geq{\mathbb{E}_{q(z|x)}[\log{p(x|z)}]-KL(q(z|x)||p(z))}\label{vae_elbo}
\end{equation}
while both \begin{math}q(z|x)\end{math} and \begin{math}p(z|x)\end{math} are parameterized as encoder \begin{math}q_{\phi}(z|x)\end{math} and decoder \begin{math}p_{\theta}(z|x)\end{math}. It is obvious that VAE encodes the input \begin{math}x\end{math} into a probability distribution rather than a fixed vector so that different \begin{math}z\end{math} could be chosen from the distribution to obtain different outputs \begin{math}x\end{math}.

The VAE model could be modified to be conditioned on a certain attribute \begin{math}c\end{math} such as dialogue contexts to generate outputs given a source pattern. And such modification leads the original VAE to conditional VAE called CVAE~\cite{yan2016attribute2image,sohn2015learning}. Needless to say, the output of CVAE now depends both on \begin{math}z\end{math} and \begin{math}c\end{math} and the ELBO becomes:
\begin{gather}
{\log{p(x|c)}}\nonumber \\
\geq{\mathbb{E}_{q(z|x,c)}[\log{p(x|z,c)}]-KL(q(z|x,c)||p(z|c))}\label{cvae_elbo}
\end{gather}

VAE and CVAE seem to show great potential to generate diverse outputs, as the latent variable \begin{math}z\end{math} from a distribution could be modulated to help model different patterns. Some image generating tasks adopt VAE or CVAE and achieve good generative quality. In spite of this, difficulties are encountered while researchers attempted to generate texts via such structures~\cite{bowman2016generating}. Directly optimized with Equation \ref{vae_elbo} or Equation \ref{cvae_elbo} will lead to the \emph{KL-Vanishing problem} which is also called as the \emph{posterior collapse problem}. For the VAE model, the encoder \begin{math}q_{\phi}(z|x)\end{math} perfectly fits the prior distribution \begin{math}p(z)\end{math} while ignores the inputs and the decoder generates outputs without referring to \begin{math}z\end{math}. Same problems take place all conditioned on \begin{math}c\end{math} while CVAE is used.

\cite{bowman2016generating} presented the \emph{KL annealing} method (KLA) and \emph{word-dropout} operation (WD) in VAE training to mitigate the \emph{KL-Vanishing problem} in their sentence generating system. And \cite{zhao2017learning} introduced an additional \emph{bag-of-word loss} which takes the latent variable as input and predicts the words which will appear in the target so that the connection of the latent variable and the outputs are augmented. They applied their \emph{bag-of-word} loss in a CVAE structure for one-to-many text generation tasks and get excellent results in open domain dialogue generation of discourse-level diversity.

\section{The Proposed Method}
\subsection{Analysis of the KL-Vanishing Problem}
Considering VAE's objective function Equation.~\ref{vae_elbo}, two facts are observed:
(1) The second term \begin{math}KL(q(z|x)||p(z))\end{math} reaches its global minimum of 0 when \begin{math}q(z|x)=p(z)\end{math}. 
(2) According to Jensen's Inequality, \begin{math}E_{p(z)}[\log{p(x|z)}]\leq\log\sum_{z}[p(x|z)p(z)]=\log{p(x)}\end{math}, and the equal sign of the inequation holds when and only when \begin{math}x\end{math} is independent of \begin{math}z\end{math}.
As a consequence, when $x$ and $z$ are independent, the ELBO objective degenerates to original $\log p(x)$ objective under which the decoder learns a plain language model.
Thus the encoder $q(z|x) = p(z)$ and decoder $p(x|z) = p(x)$ which fits the dataset as a plain language model constitute a trivial solution of the objective of VAE.
At this time, the KL divergence term in ELBO becomes 0, and we call this phenomenon \emph{KL-Vanishing}.

Although the \emph{KL-Vanishing} point is a solution of Equation. \ref{vae_elbo}, it is not actually we want.
When \emph{KL-Vanishing} takes place, Equation. \ref{vae_elbo} degenerates to \begin{math}\log{p(x)}\end{math}.
 When the decoder is modeled by an RNN structure for text generation, it easily converges to a average language model of target sentences without regarding to $z$ under the degenerated objective.
 As a result, \begin{math}z\end{math} losses its ability of affecting \begin{math}x\end{math} and the decoder fits the average behavior of all \begin{math}x\end{math}.
 
 \begin{figure}[t]
	\centering  
	\subfigure[] { \label{fig:distinguishable} 
	\includegraphics[width=0.43\columnwidth]{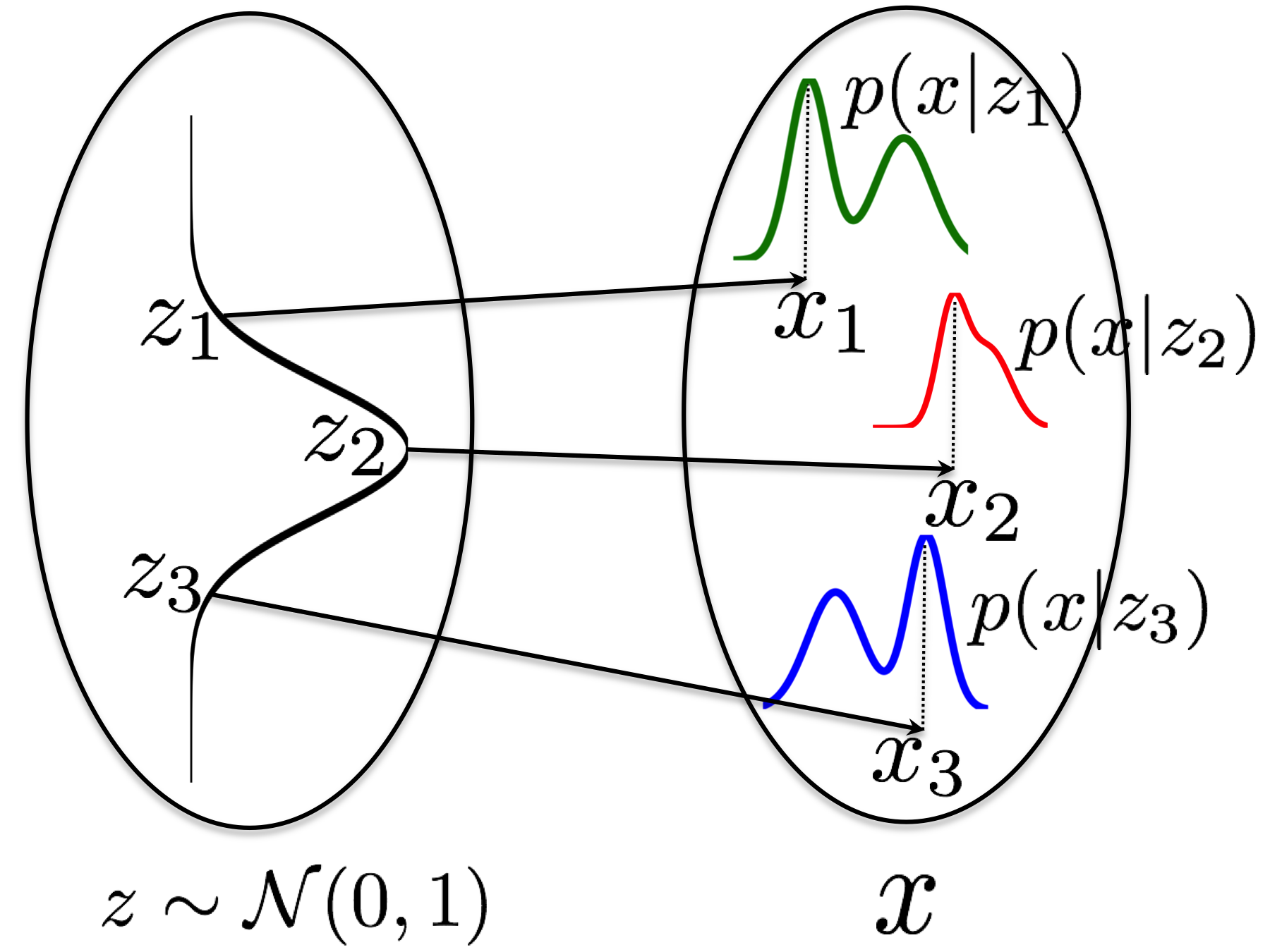}}
	\subfigure[] { \label{fig:kl-vanishing}
	\includegraphics[width=0.43\columnwidth]{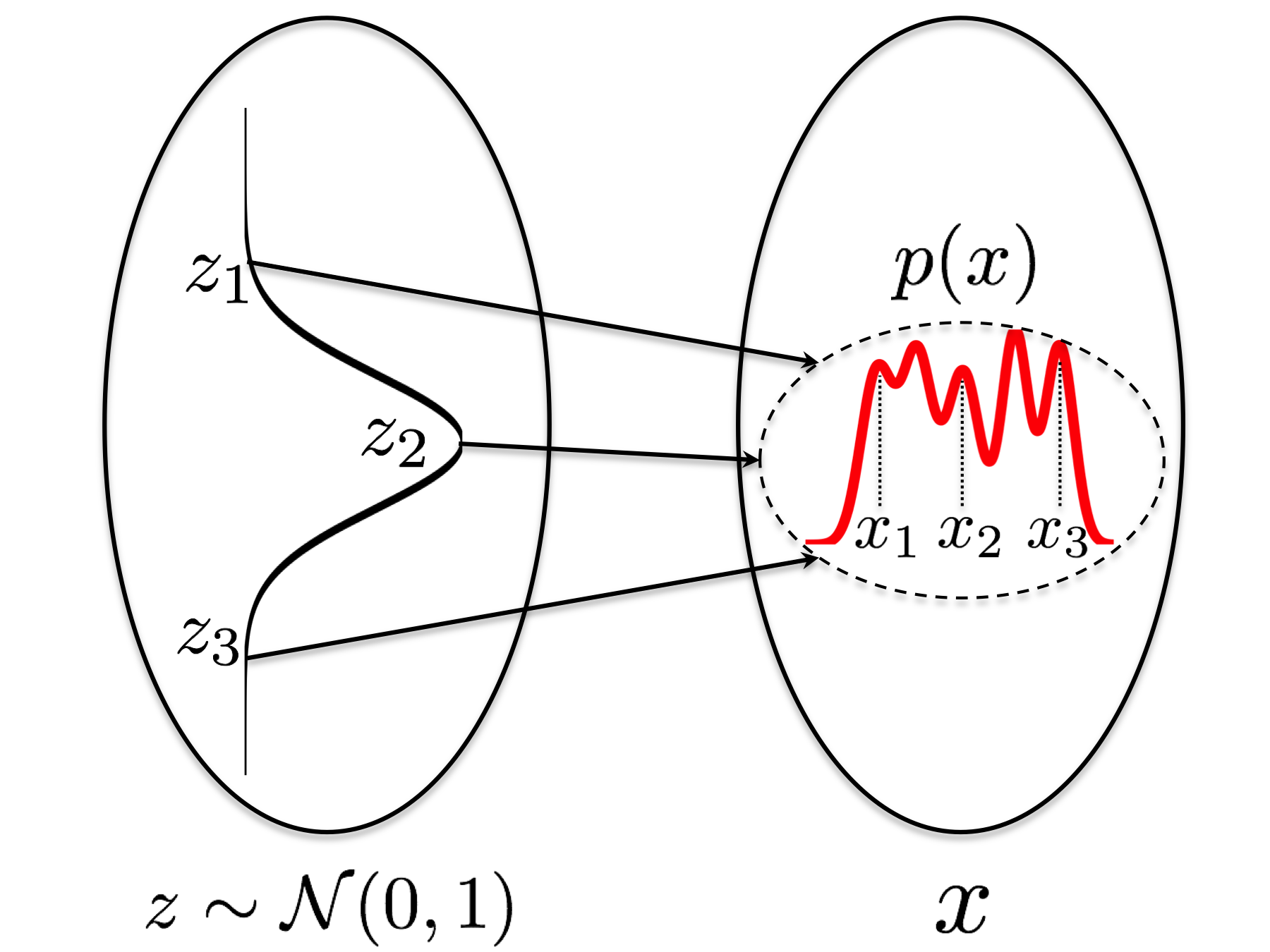}}
	\vspace{-0.2cm}
	\caption{Illustration of the generation process and KL-vanishing. (a) In assumption, each configuration of the latent variable is mapped by the decoder into a different decoding distribution $p(x|z)$. Benefiting from the latent distribution of $z$, all $p(x|z)$ with different $z$ can have a good coverage of the entire target space of $x$, while $p(x|z)$ itself can be simple and only responsible for decoding one target. (b) When KL-vanishing takes place, $z$s loss the expressiveness of $x$ and collapse to a same decoding distribution $p(x)$. Thus $p(x)$  trying to fit the entire space of $x$ alone has a very complex structure and might only have a poor coverage of the space and lack the diversity.}  
	\vspace{-0.4cm}
	\label{fig:latentspace}   
\end{figure}

Previous works tried to avoid the \emph{KL-Vanishing problem} by either breaking fact (1) or fact (2). Some of them put efforts into preventing \begin{math}q(z|x)\end{math} from collapsing to \begin{math}p(z)\end{math} by slowing down the optimizing procedure of the KL term.
Others attempt to force the decoder to depends on the \begin{math}z\end{math} by weakening the decoder or making encoder more complicated.
However, such methods fall into a dilemma: On the one hand, the objective function is optimized to reach its maximal. On the other hand, the optimal point should be avoided to prevent the occurrence of \emph{KL-Vanishing}.
These methods tried to find a good balance between the two contradictory facts without an explicit objective.
Nevertheless, it is hard to find an appropriate trade-off point.

\subsection{Self Labeling CVAE}
Our goal is to generate diverse \begin{math}x\end{math} using different \begin{math}z\end{math}.
From the perspective of the decoder, two conditions should be satisfied: First, each \begin{math}z\end{math} should correspond to a unique \begin{math}x\end{math} through the decoder.
Second, \begin{math}z\end{math} should obey the prior distribution \begin{math}p(z)\end{math}.
Maximizing the ELBO objective encourages the latter by pulling encoder's output distribution of \begin{math}z\end{math} to \begin{math}p(z)\end{math}.
However, with the $p(z|x)$ moving towards $p(z)$ during the optimizing procedure, $z$ losses the discriminative information and the decoder tends to fit the data even without the help of encoders.
As a consequence, the first condition is violated and multiple \begin{math}z\end{math}s will collapse to a same averaged output distribution \begin{math}p(x)\end{math}, as is shown in Fig.~\ref{fig:latentspace}(b).

Thus we propose to strengthen the connection between the latent $z$ and target $x$ via maintaining the expressiveness of the encoder.
As illustrated in Fig.~\ref{fig:latentspace}(a), considering that an expressive $z$ has the ability to recover a unique target through the decoder, the decoder itself can then be used to find the most expressive $z'$ given a certain target $x$. 
This in concept equivalents to find the inverse image of $x$ of the decoder.
So the inverse image $z'$ of $x$ can be regarded as the effectiveness label $x$ in the continuous latent space.
And if $z'$ has been obtained, then we are reasonably motivated to pull the encoder distribution $p(z|x)$ to close $z'$ to maintain the expressiveness of the encoder.

However, finding the inverse image of the decoder exactly is not an easy task.
To overcome this, we introduce an extra network to estimate the inverse image of $x$ of the decoder, i.e. the effectiveness label.
Thus we call this network the \emph{labeling network} and denote the output of it as $z_{label}$.
The \emph{labeling network} can also be considered as an approximation to the ideal encoder for current decoder.
Specifically, the \emph{labeling network} shares the same network structure with the original encoder of VAE, but it only outputs the variable $z_{label}$ rather than the reparameterized distribution.
As the output of the VAE encoder is a distribution $q(z|x)$, we put the expressive constraint on the expectation of the $L_2$ distance ${||z-z_{label}||}^2$ between encoded latent variable and $z_{label}$ over the encoder distribution $q(z|x)$.
Thus an expressiveness objective function is defined as follows:
\begin{equation}
\mathcal{L}_\mathsf{exp} = \mathbb{E}_{q(z|x)}[{||z-z_{label}||}^2]\label{eq:vae_pre}
\end{equation}
which is minimized to encourage the encoder to be more expressive.
By using $g(x)$ to denote the labeling network i.e. $z_{label}=g(x)$, and adding $\mathcal{L}_\mathsf{exp}$ as an additional term to the VAE's objective function, we get the total objective function in following:
\begin{gather}
\mathcal{L}_\mathsf{SLVAE} = -\mathbb{E}_{q(z|x)}[\log{p(x|z)}]+KL(q(z|x)||p(z))\nonumber \\
+\lambda\mathbb{E}_{q(z|x)}[{||z-g(x)||}^2]\label{eq:slvae}
\end{gather}
From this formulation, we can see that, $q(z|x)$ is not only pulling to $p(z)$ like before, but also pulling to the estimated ``best encoder'' for the decoder.
The hyper-parameter $\lambda$ is used to control the importance of the expressiveness objective.
The ``best encoder'' can expand a comprehensive coverage of the target space through the current decoder.
Thus they will reach an equilibrium at which the $p(z|x)$ is close to $p(z)$ and also remains the expressiveness.
As we use a labeling network to estimate the most expressive latent label given the decoder and strengthen the connection between the latent $z$ and target $x$ through the decoder itself, we call this method Self Labeling VAE~(SLVAE).

When  it comes to CVAE, things remain the same except that everything is conditioned on \begin{math}c\end{math}.
And the final total objective function is:
\begin{gather}
\mathcal{L}_\mathsf{SLCVAE} = -\mathbb{E}_{q(z|x,c)}[\log{p(x|z,c)}]+KL(q(z|x,c)||p(z|c))\nonumber \\
+\lambda\mathbb{E}_{q(z|x,c)}[{||z-g(x,c)||}^2]\label{eq:slcvae}
\end{gather}
Similarly, we call this model SLCVAE.

At last, learning the \emph{labeling network} $g(x, c)$ is somewhat straightforward.
As we discussed before, \begin{math}g(x,c)\end{math} should be the ``best encoder" for the decoder to recover \begin{math}x\end{math}.
Thus we should optimize \begin{math}g(x,c)\end{math} by maximizing the following objective function:
\begin{gather}
\log{p(x|z_{label},c)}=\log{p(x|g(x,c),c)}\label{eq:labeling}
\end{gather}
with the decoder fixed.
An alternative training schedule of the VAE/CVAE network and the labeling network is applied.

Fig.~\ref{fig:slcvae} shows the overview of the whole proposed method.
The CVAE part of our structure is a conventional CVAE with reparameterization trick\cite{kingma2013auto} by introducing a posterior and a prior network as described in \cite{zhao2017learning} in detail.
The structure of the labeling network is similar to the encoder of the CVAE which consists of a target encoder and a context encoder which embed target and source sentences to representative vectors.
The only difference is that the labeling encodes its inputs into a fixed \begin{math}z_{label}\end{math} rather than a distribution.

\subsection{Training Process}
To optimize Equation.~\ref{eq:slcvae} and Equation.~\ref{eq:labeling}, we parameterize all the three modules: the encoder \begin{math}q_{\phi}(z|x, c)\end{math} and decoder \begin{math}p_{\theta}(x|z, c)\end{math} of the CVAE, and the labeling network \begin{math}g_{\gamma}(x, c)\end{math}.
An alternative training schedule is used with two phases: the CVAE phase and the Labeling phase.

In the CVAE phase, we minimize the loss function of the SLCVAE:
\begin{gather}
\min_{\phi,\theta,\beta}\mathcal{L}_\mathsf{SLCVAE}=\min_{\phi,\theta,\beta}[-\mathbb{E}_{q_{\phi}(z|x,c)}[p_{\theta}(x|z,c)]\nonumber \\
+KL(q_{\phi}(z|x,c)||p_{\beta}(z|c)) \\
+\lambda\mathbb{E}_{q_{\phi}(z|x,c)}[{||z-g_{\gamma}(x,c)||^2}]]\nonumber\label{eq:lslcvae}
\end{gather}
where \begin{math}\beta\end{math} are parameters of the prior network.
In this phase, the labelling network $g_{\gamma}(x,c)$ is fixed to provide a $z_{label}$ corresponding to each $x$.

In the Labeling phase, we minimize the loss function of the labeling network:
\begin{gather}
\min_{\gamma}\mathcal{L}_\mathsf{label}=\min_{\gamma}[-p_{\theta}(x|g_{\gamma}(x,c), c)]\label{eq:llabel}
\end{gather}
The decoder is fixed at this time to get the good expressive label for current decoder.

See Algorithm~\ref{alg:slcvae} for the whole training procedure.
The Adam\cite{kingma2015adam} optimizer is adopted to update parameters.

\begin{algorithm}[t]
\caption{Alternative training procedure of SLCVAE. We fixed $m=n=1$ in all our experiments to speedup the training.}
\label{alg:slcvae} 
\begin{algorithmic}[1]
\STATE Initialize ${\phi}, {\theta}, {\beta}, {\gamma}$ randomly
\FOR{number of iterations}
	\FOR{m steps}
		\STATE $c, x \gets$ sample a mini-batch from dataset
		\STATE Sample latent $z \sim q_{\phi}(z|x, c)$
		\STATE Calculate label $z_{label} \gets g_{\gamma}(x,c)$
		\STATE Calculate the gradients: $\nabla_{\phi,\theta,\beta}\mathcal{L}_\mathsf{SLCVAE}$
		\STATE Update CVAE parameters $\theta, \phi, \beta$ by Adam
	\ENDFOR
	\FOR{n steps}
		\STATE $c, x \gets$ sample a mini-batch from dataset
		\STATE Calculate the gradients: $\nabla_{\gamma}\mathcal{L}_\mathsf{label}$
		\STATE Update labeling network parameter $\gamma$ by Adam
	\ENDFOR
\ENDFOR
\end{algorithmic}
\end{algorithm}

\section{The EGOODS Dataset}
The text generating problem defined with ``one source, multiple targets'' is an active research topic and plays important roles in many tasks.
However, there still lacks real one-to-many datasets to improve and evaluate the algorithms for this problem.
Most current datasets come from dialogue system are essentially one-to-one corpora.
Although there may exist various underlying responses for a certain question, these datasets only contain one answer for each dialogue context due to data source limitations.

To fulfill the gap between the demand and status quo for one-to-many dataset, we collect a large scale item description corpus from a Chinese e-commerce website to construct the native one-to-many dataset.
In this corpus, each item has one description provided by their sellers and multiple recommendation sentences written by third-party who is payed to make these sentences more attractive to customers.
The descriptions provided by sellers are usually texts stacking many keywords of the item properties.
In the contrary, the recommendation sentences are written according to item descriptions but read more smoothly.
For the text generation task, we naturally use the seller's descriptions as the source to generate multiple recommendation sentences mimicking humans.
Since this corpus originates from a real business, texts are of high quality and coherent with sources.
Thus it gives rise to a very large and native one-to-many dataset, which is called EGOODS.

After simple cleaning and formatting, EGOODS dataset contains 3001140 source and target pairs from 789582 items in total.
So each source item description has 3.8 target recommendation sentences on average.
The dataset is split into training/validation/testing parts with respect to items, each of which contains 2961317/19536/20287 pairs.

\section{Experiments} 
\subsection{Experimental Setups}
\subsubsection{Datasets}
Our experiments are conducted on two text generating tasks: open-domain dialogue generation and recommendation sentence generation.
We evaluate the performance of our algorithm and compare it with several strong baselines on the two tasks.
For the first task,  the public dialogue dataset Daily Dialog (DD) \cite{li2017dailydialog} is used.
DD dataset is collected from different websites under 10 topics.
It contains 13118 multi-turn dialogue sessions in English, and is split into training/validation/testing set of 11118/1000/1000 sessions.
The average number of turns per session is 8.85 and the average number of tokens per utterance is 13.85.
To avoid too long dialogue contexts, we first split long sessions into multiple short full speaker turns containing no more than 6 utterances with an utterance level sliding window.
As a result, the final DD dataset contains 39567/3681/3471 full speaker turns in training/validation/testing set.
For each full speaker turn, we use all utterances but the last one as the dialogue context to predict the last one.
Need to note that though there may exist various responses for a question, DD dataset essentially only contains one-to-one data.
To better model and evaluate the diversity, the newly constructed one-to-many dataset EGOODS is adopted in the second task.

\subsubsection{Baselines}
We compared our SLCVAE to 4 strong baselines: SEQ2SEQ~\cite{sutskever2014sequence}, MMI-AntiLM~\cite{li2016diversity}, CVAE and CVAE with \emph{bag-of-word loss} (CVAE+BOW)~\cite{zhao2017learning}.
Several training skills, such as \emph{KL-annealing}(KLA) and \emph{word dropout}(WD)~\cite{bowman2016generating}, are used in combination with baselines and our method to improve the performance.
All methods are required to generate 10 responses for each given input.

Note that although the SEQ2SEQ model uses deterministic encoding vectors, the widely adopted beam search strategy can be applied during inference procedure to generate 10-best decoding results which corresponds to 10 responses~(denoted as SEQ2SEQ-BS).
MMI-AntiLM method follows this idea and put MMI prior onto the beam search strategy.
We tried different beam size from 10 to 100 and find that beam size set to 10 gives the best result for all dataset.
It is worth noting that beam search is applied only to above two methods and will bring unfair advantages due to the exploration of a much larger search space.
All other methods including ours use the greedy strategy during decoding to be consistent with previous work.
We also tried another simple strategy that adds a random noise drawn from a gaussian distribution to the encoded vector to bring variability to SEQ2SEQ.
We denote this method as SEQ2SEQ + noise, and use it as extra baselines with various standard deviations.




\subsubsection{Training}
The whole structure of SLCVAE is implemented with the famous open source library PyTorch\cite{paszke2017automatic}.
In all experiments, English letters are all transformed to the lower case first.
Encoders are bidirectional RNNs~\cite{schuster1997bidirectional} with Gated Recurrent Units (GRU)~\cite{chung2014empirical} and the decoders are unidirectional RNN with GRUs throughout all experiments.
All RNNs have two layers.
Since EGOODS dataset is much larger than DD, the network capacity increases accordingly.
Specifically, for DD and EGOODS dataset respectively, the word embedding sizes are set to 32 and 128, and the hidden dimensions of RNN are also set as the same.
In all VAE-based methods, the latent variable dimensions are set to 8 and 16 for two datasets separately.
The Adam optimizer with a learning rate of 0.0001 is used to train all models with batch sizes of 64 and 128 for two datasets.
Training skills of KLA and WD are also used to get further better performance.
We also conduct an annealing strategy that the weight of our labeling error is increased over time synchronously with KLA, as the soft label provided by the labeling network is not that good in the early stages.
We tune several hyper-parameters on the validation sets and measure the performances on the test sets for all baselines and our proposed method.

\subsection{Results}
\subsubsection{Evaluation Metrics}
In ``one source, multiple targets'' setting, for an input $c$, given $N$ hypothesis responses $h_i$ generated by a model and $M_c$ reference responses $r_j$, accuracy and diversity are two sides of the generations we need to concern.
Automatic quantitative measures for these purposes are still an open research challenge~\cite{liunot,tong2018one}.
\cite{zhao2017learning} proposed BLEU-precision and BLEU-recall metrics for discourse-level accuracy and diversity respectively as following:
\begin{equation}
\begin{split}
\mathsf{precision(c)} &= \frac{\sum_{i=1}^{N}\max_{j \in [1,M_c]}d(r_j, h_i)}{N} \\
\mathsf{recall(c)} &= \frac{\sum_{j=1}^{M_c}\max_{i \in [1,N]}d(r_j, h_i)}{M_c}
\end{split}
\end{equation}
BLEU-1, BLEU-2 and BLEU-3 are adopted and their average is calculated as the metrics.
However, BLEU-recall is defined based on lexical similarity, which might penalize a reasonable but not same prediction.
Following \cite{li2016diversity}, we also use the number of \emph{distinct} \emph{n-gram} to measure the word-level diversity.
The \emph{distinct} is normalized to [0, 1] by dividing the total number of generated tokens.
In summary, BLEU-precision is reported as the accuracy measure, and BLEU-recall,  \emph{distinct-1} and \emph{distinct-2} are reported as diversity measures.

\begin{figure}[t]
	\centering  
	\includegraphics[width=0.95\linewidth]{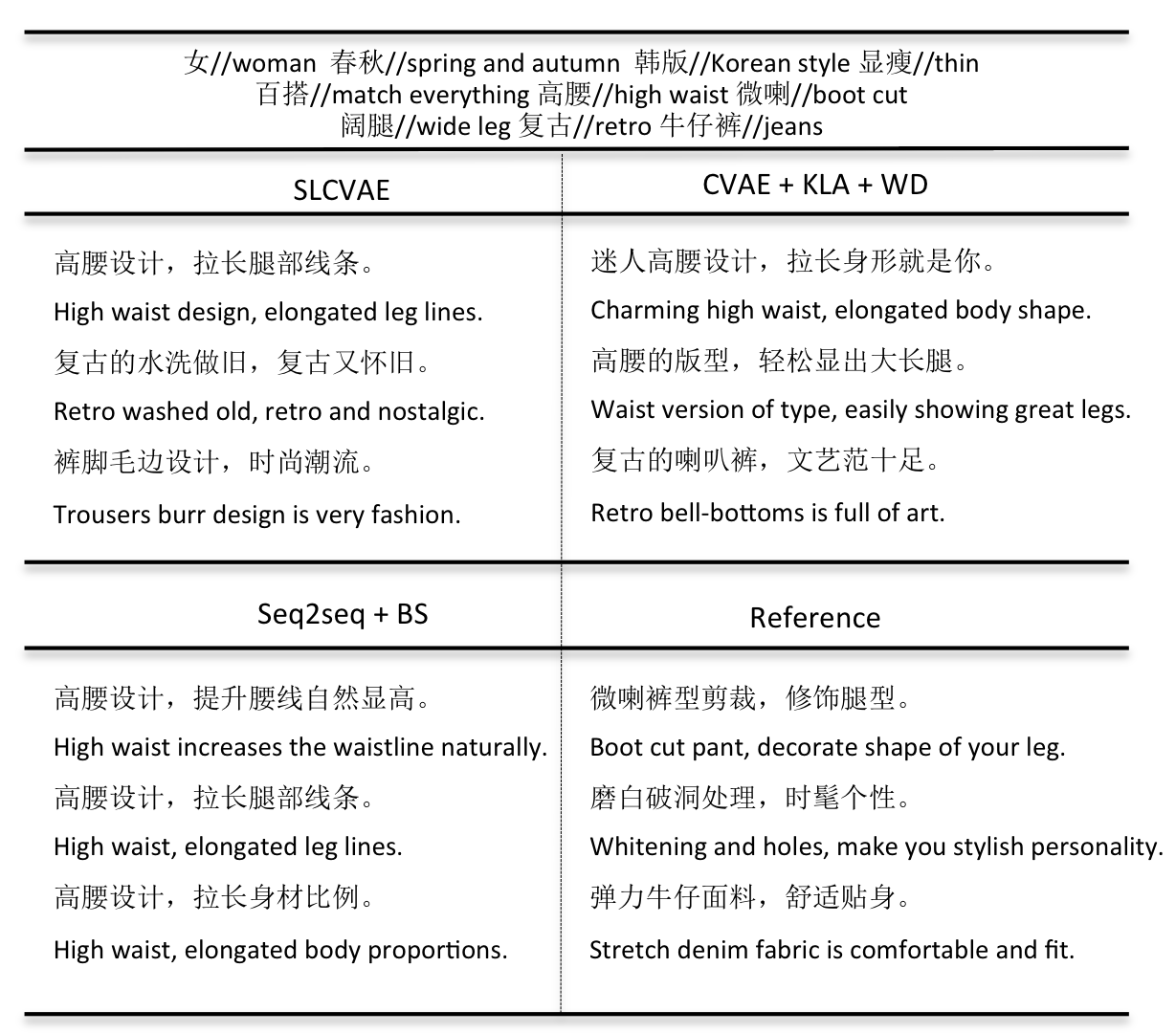}  
	\vspace{-0.4cm}
	\caption{Example of generated texts. Despite similar coverage for references, SLCAVE has better diversity in vocabulary and expressions.}  
	\vspace{-0.4cm}
	\label{fig:show_case}   
\end{figure}
We also conduct human evaluations on the EGOODS dataset. 7 human experts are employed to measure the fluency of generated sentences, coherence of each sentence to source and diversity. For fluency and coherence, experts are asked to vote to each sentence. Sentences which yield more than 4 votes are good sentences. The ratio of good sentences are reported. For diversity,  5 level of diverse scores are introduced. The higher the score, the more diverse the sentence is. The final diversity score of each sentence is the average score of all experts.

\subsubsection{Automatic Quantitative Measurement on Daily Dialog}
Table.~\ref{tab:dd_result} shows the evaluation results of all methods on Daily Dialog dataset.
Results for SEQ2SEQ-noise with different standard deviations are also listed.
Training skills of KLA and WD are used for all CVAE based methods.
We can see that our proposed method outperforms all baselines in terms of all the 4 metrics on this task.
And it is worth noting that our method obtains much higher diversity measures no matter in discourse-level or word-level than all others.
In the meanwhile, the accuracy metrics BLEU-precision of our method remains slightly better than the best baseline.
This confirms our insight of the generating process that our labeling objective can lead to an equilibrium at which the KL-vanishing problem is significantly relieved and so result in better diversity.
Remind that Daily Dialog is actually a \emph{one-to-one} dataset.
The better performance in diversity on DD demonstrates that our model can better exploit such training data without explicit \emph{one-to-many} annotations.
Further, the results show that CVAE based models beat conventional encoder-decoder methods in almost all metrics.
This is consistent with those in previous work like~\cite{zhao2017learning}, and confirms the advantage of latent variable methods for generation tasks over encoder-decoder models with multi-decoding strategy.
In addition, we find that with growing noise, the accuracy of SEQ2SEQ+noise decreases while the diversity increases significantly.
Especially when the noise is small (e.g. 0.2), the diversity has obvious gains with only a slightly sacrifice on accuracy.

\begin{table}
	\caption{Results on Daily Dialog (DD). The bottom 3 lines are CVAE based methods. }
	\label{tab:dd_result}
	\begin{tabular}{lp{0.85cm}p{0.85cm}p{0.85cm}p{0.85cm}}
		\toprule
		Methods & BLEU-prec & BLEU-recall & \emph{distinct-1} & \emph{distinct-2}\\
		\midrule
		SEQ2SEQ+BS & 0.164 & 0.282 & 0.002 & 0.007\\
		SEQ2SEQ+noise(0.2) & 0.163 & 0.288 & 0.003 & 0.014\\
		SEQ2SEQ+noise(0.5) & 0.157 & 0.312 & 0.005 & 0.032\\
		SEQ2SEQ+noise(0.8) & 0.153 & 0.320 & 0.007 & 0.065\\
		MMI-AntiLM & 0.153 & 0.275 & 0.002 & 0.012\\
		\midrule
		KLA+WD & 0.212 & 0.345 & 0.010 & 0.041 \\
		KLA+WD+BOW & 0.210 & 0.344 & 0.013 & 0.066\\
		\midrule
		KLA+WD+SL & \textbf{0.214} & \textbf{0.354} & \textbf{0.014} & \textbf{0.078} \\
		\bottomrule
	\end{tabular}
	\vspace{-0.4cm}
\end{table}

\subsubsection{Automatic Quantitative Measurement on EGOODS}
To better study current methods on the ``one source, multiple targets'' problem, experiments are conducted on our newly collected native \emph{one-to-many} dataset EGOODS.
Performances of different methods are shown in Table.~\ref{tab:egoods_result}.
First of all, our method achieves comparable accuracy with baselines and best diversity among all methods with an only exception of SEQ2SEQ+noise(0.8) on \emph{distinct-2}.
This demonstrates the effectiveness of SLCVAE on the \emph{one-to-many} data.
Although SEQ2SEQ+noise(0.8) gets the best \emph{distinct-2}, its precision is sacrificed significantly due to the nosie, which means the results tend to be meaningless.
In detail, our method harvests the much better gains on word-level diversity while is only slightly better than CVAE on BLEU-recall.
We explain this in two folds:
First, strong baselines can benefit from the large scale and \emph{one-to-many} nature of EGOODS to better fit the multiple targets. 
Another reason is that automatically evaluating the quality of generated texts is very challenging.
BLEU-recall only measures the coverage of hypothesis for the annotated targets, and could not judge good algorithms precisely when the annotations are limited.
In such situation, \emph{distinct} measures the vocabulary a model actually uses and demonstrates its absolute lexical diversity.
Example results will show this in next subsection. 

Furthermore, we observed that SEQ2SEQ+BS obtains the best BLEU-precision among all methods on EGOODS, but it performs much worse on Daily Dialog.
Meanwhile, the BLEU-recall gap between SEQ2SEQ+BS and the best result on EGOODS is obviously small than that on DD.
We point out that our dataset especially designed for ``one source, multiple targets'' problem significantly improves the generation quality of SEQ2SEQ methods.


\begin{table}
	\caption{Results on EGOODS. The bottom 3 lines are CVAE based methods.}
	\label{tab:egoods_result}
	\begin{tabular}{lp{0.85cm}p{0.85cm}p{0.85cm}p{0.85cm}}
		\toprule
		Methods & BLEU-prec & BLEU-recall & \emph{distinct-1} & \emph{distinct-2}\\
		\midrule
		SEQ2SEQ+BS & \textbf{0.379} & 0.388 & 0.0012 & 0.0042\\
		SEQ2SEQ+noise(0.2) & \textbf{0.379} & 0.386 & 0.0021 & 0.0146\\
		SEQ2SEQ+noise(0.5) & 0.367 & 0.402 & 0.0029 & 0.0162\\
		SEQ2SEQ+noise(0.8) & 0.347 & 0.395 & 0.003 & \textbf{0.0420}\\
		MMI-AntiLM & 0.356 & 0.374 & 0.0021 & 0.0146\\
		\midrule
		KLA+WD & 0.373 & \textbf{0.405} & 0.0039 & 0.0216\\
		KLA+WD+BOW & 0.374 & 0.404 & 0.0039 & 0.0231\\
		\midrule
		KLA+WD+SL & 0.373 & \textbf{0.405} & \textbf{0.0049} & 0.0270 \\
		\bottomrule
	\end{tabular}
	\vspace{-0.4cm}
\end{table}



\subsubsection{Human Evaluation on EGOODS}
Human evaluation results on EGOODS are shown in Table~\ref{tab:human_eval}.
Such results show that our method achieves comparable fluency and coherence as baseline methods, but our diversity is much higher than other models.
\begin{table}[h]
\vspace{0cm}
\small
\renewcommand\arraystretch{1}
	\caption{Human evaluation results.}
	\label{tab:human_eval}
	\begin{tabular}{lp{1.3cm}p{1.8cm}p{1.2cm}}
		\toprule
		Methods & Fluency(\%) & Coherence(\%) & Diversity \\
		\midrule
		SEQ2SEQ+BS & \textbf{96} & 65 & 1.55 \\
		KLA+WD & 87 & 64 & 3.12 \\
		KLA+WD+BOW & 83 & \textbf{66} & 3.18 \\
		\midrule
		KLA+WD+SL & 91 & \textbf{66} & \textbf{3.32} \\
		\bottomrule
	\end{tabular}
	\vspace{0cm}
\end{table}
Although the SEQ2SEQ+BS method achieves the best fluency, it sacrifices too much diversity, which means the result is monotonous and dull.

\subsubsection{Text Generating Examples}
To give an intuitive impression about generations,  Fig.~\ref{fig:show_case} shows an example of generated texts for EGOODS, and more results can be found in the supplementary material.
10 results are generated separately by SEQ2SEQ+BS, CVAE and our method SLCVAE.
The results from all three methods are of good fluency and coherent to the input.
But obviously SEQ2SEQ+BS fails to show different expressions thus gets poor diversity.
Both CVAE model and our method tend to show stronger abilities in generating diversely than SEQ2SEQ+BS, since we can see that the generated results have better coverages for the references.
Nevertheless, notice that the SLCVAE has a larger vocabulary and uses richer expressions that CAVE+BOW which is not reflected by BLEU-recall metrics.
This finding is consistent with the quantitative experiment results we have discussed above.


\subsection{Conclusion}
``One source, multiple targets'' is a common text generation task.
Recently CVAE based methods shows great potentials for this task.
However CVAE working with RNNs tends to run into the KL-vanishing problem that the RNN ends up with a trivial language model independent of the latent variable.
In this paper, we analyze the objective of CVAE and give an intuitive explanation of the cause of KL-vanishing.
Then we propose the self labeling mechanism which connects the decoder with latent variable by an explicit optimization objective.
It leads the encoder to reach an equilibrium at which the decoder can take full advantage of the latent variable.
Experiments show that SLCAVE largely improves the generating diversity.

\bibliography{Bibliography-File}
\bibliographystyle{aaai}
\end{document}